\documentclass[11pt]{article}

\usepackage[preprint]{acl}
\usepackage{times}
\usepackage{latexsym}
\usepackage[T1]{fontenc}
\usepackage[utf8]{inputenc}
\usepackage{microtype}
\usepackage{inconsolata}
\usepackage{graphicx}
\usepackage{booktabs}
\usepackage{array}
\usepackage{amsmath}
\usepackage{url}

\newcommand{\ins}[1]{\texttt{#1}}

\title{When ``Neutral'' Means Irrelevant: A Construct-Validity Audit of NormAd-ETI}

\makeatletter
\newcommand{\eqcontrib}{%
  \begingroup
    \renewcommand{\@makefnmark}{}%
    \thanks{Equal Contribution.}%
  \endgroup}
\makeatother

\author{\normalfont
  Aarav Singh\textsuperscript{1,*}\eqcontrib
  \quad Animesh Pathak\textsuperscript{1,*}
  \quad Christian Hoang\textsuperscript{2,*} \\
  Navyansh Singh\textsuperscript{1}
  \quad Avantika Singh\textsuperscript{1} \\[2pt]
  {\normalfont\small \textsuperscript{1}IIIT Naya Raipur
  \qquad \textsuperscript{2}FPT University}%
}

\begin{document}
\maketitle

\begin{abstract}
NormAd-ETI scores a model's judgment of everyday actions as acceptable, unacceptable, or neither as the cultural context shrinks from a specific rule to a country name. We audit what that three-way score measures. It combines two decisions---whether a supplied norm applies, and whether the action conforms to it. Separating them shows that when only a country is shown, applicability falls to near chance while conformity stays well above it, because the prompt withholds the rule a Neutral label depends on. Because one answer carries both decisions, the score follows the steadier conformity and hides the weaker applicability, and no single prompt recovers the two separately. Dedicated forced-choice probes separate them for frontier models but not for small open models, where the applicability judgment still tracks the hidden acceptability label; adding a written rationale reduces that conflation without removing it. We recommend reporting the two results separately and measuring the conflation gap.
\end{abstract}

\section{Introduction}

NormAd was introduced to test whether models can adapt judgments of social acceptability as the available cultural context changes, from a specific Rule-of-Thumb (RoT), to an abstract value, to a country name \citep{rao2025normad}. NormAd-ETI operationalizes this goal with everyday stories labeled Yes, No, or Neutral.

The benchmark's headline result is easy to state and easy to misread. Accuracy falls as context is removed, and it falls fastest for the Neutral class. We ask what that fall means. The Neutral label is constructed by pairing a story with a norm from outside the story's cultural cluster, so its intended meaning is that the supplied norm \emph{does not apply}. The prompt, however, asks only whether the action is acceptable and offers ``Neither'' without saying what it means. A single answer therefore stands in for two decisions: \emph{applicability} (does the norm bear on this action?) and \emph{conformity} (if it does, does the action follow it?).

We make two claims on the same items and models. The first is about the \textbf{Country} condition: a Neutral item is built from a rule that does not apply to the story, and at the Country level that rule is never shown, so the construct depends on an input the prompt withholds. Our experiments show the shortfall lands on applicability, which drops far more than conformity. The second is about \textbf{bundling and separation}: one answer carries both decisions, so the score follows the stronger conformity and hides the weaker applicability, while scoring each decision with its own probe separates them at the frontier level but not in the five small open models. We read this conflation as consistent with a representational limit in the small models rather than a flaw in the probes.

Our contributions are:
\begin{itemize}
    \item \textbf{A construct-validity framing} of the applicability/conformity ambiguity, with the measurement assumptions made explicit \citep{cronbach1955construct,jacobs2021measurement}.
    \item \textbf{A context gap at Country applicability}: the Country deficit is dominated by a loss of applicability, which falls far more than conformity.
    \item \textbf{Conflation in the separate probes at small model size}: under a dedicated applicability probe, the apply/not-apply call still tracks the hidden acceptability label in the five small open models.
    \item \textbf{Clean separation of both capabilities at the frontier level}: two forced-choice probes separate applicability from conformity for both frontier models.
\end{itemize}

\section{Related Work}

\paragraph{Cultural-adaptability benchmarks.} NormAd \citep{rao2025normad} belongs to a family of benchmarks that test cultural competence in language models. CulturalBench \citep{chiu2024culturalbench} probes cultural knowledge with multiple-choice questions; GlobalOpinionQA \citep{durmus2023globalopinions} compares model responses to cross-national survey opinion, and other work groups countries into value clusters through the World Values Survey and the Inglehart--Welzel map \citep{inglehartwelzel2023map}; \citet{naous2024beer} probe culturally inappropriate associations. Each of these reduces cultural judgment to a single score or label. Our audit asks what the score of one such benchmark actually measures, rather than proposing another benchmark.

\paragraph{Benchmark validity and evaluation reliability.} Construct validity asks whether a measurement procedure tracks the construct its label names \citep{cronbach1955construct}, and for machine-learning systems this is a property of the measurement model encoded in the data and the prompt \citep{jacobs2021measurement}. Recent work questions the reliability of cultural-alignment evaluation: \citet{khan2025randomness} show that decoding and small prompt changes can outweigh any real cross-cultural signal, and \citet{sorensen2024roadmap} argue that evaluation should represent more than one value system. Our concern is complementary and orthogonal to decoding noise: with decoding held fixed, one condition (Country) asks a question the prompt leaves underdetermined. NormAd is also reused in larger multi-task studies of cultural alignment---\citet{agarwal2026fluent} compare Indian-focused and general-purpose models across several tasks including NormAd.

\paragraph{Social norms and acceptability.} Datasets such as Social Chemistry 101 \citep{forbes2021social}, Social IQa \citep{sap2019socialiqa}, and ETHICS \citep{hendrycks2021ethics} annotate social or moral judgments over everyday situations. NormAd's Neutral label instead turns on relevance rather than normative ambiguity. Separating relevance from conformity is a measurement distinction that other social-judgment datasets may also blur, but we test it here only on NormAd.

\paragraph{Decomposing judgments and eliciting rationales.} Chain-of-thought prompting \citep{wei2022cot} and self-consistency \citep{wang2023selfconsistency} elicit intermediate reasoning, but such explanations need not be faithful: \citet{turpin2023unfaithful} show models rationalize biased answers, while models can also report calibrated uncertainty in words \citep{kadavath2022know,lin2022uncertainty}. Our two forced-choice probes are not a reasoning method but a measurement decomposition; and our paired rationale runs ask whether writing a rationale before the answer changes the labels, a question distinct from whether the rationale is faithful.

\paragraph{Metrics for two-class evaluation.} Isolating one decision makes class imbalance and degenerate response margins central. We use Youden's $J$ \citep{youden1950index} rather than the Matthews correlation coefficient \citep{matthews1975comparison,gorodkin2004comparing,chicco2020advantages}, because $J$ does not renormalize by the predicted-positive marginal and so cannot be inflated by a skewed prediction rate; paired comparisons use the exact McNemar test \citep{mcnemar1947note} with Benjamini--Hochberg correction \citep{benjamini1995controlling}.

\section{NormAd-ETI as a Measurement Instrument}

\paragraph{Dataset.} NormAd-ETI contains 2,615 everyday-etiquette stories from 75 countries, each labeled Yes, No, or Neither: 943 Yes, 875 No, and 797 Neutral items, so 1,818 (69.5\%) are applicable and 797 (30.5\%) are Neutral. Every story is presented at three levels of context: a Rule-of-Thumb (RoT), a country plus an abstract value (CVAL), or a country alone (Country). A Neutral item is one whose paired rule comes from outside the story's own cultural cluster \citep{inglehartwelzel2023map}, so the label is meant to say that the rule does not apply. The model answers a single question with the three labels, and ``Neither'' does not say whether it means ``does not apply'' or ``is ambiguous.''

\paragraph{Why the distinction matters.} Construct validity asks whether a score supports the interpretation attached to it \citep{cronbach1955construct}; for computational systems the measurement model should be explicit and tested \citep{jacobs2021measurement}. The ternary score combines two decisions: applicability (does the norm bear on the action?) and conformity (does the action follow it?). A Neither given because the norm could not be found does not show that an ambiguous case was resolved, and a NotApply given because the action seems wrong contaminates applicability with conformity. We therefore score the two decisions separately and report the conflation gap alongside.

\section{Experimental Design}
\label{sec:design}

\subsection{Data and models}

We use the corrected \texttt{normad\_v2} release ($N=2{,}615$), distributed by the benchmark's own authors in a later update to the dataset and paper. It regenerates 219 of the 797 gold-Neutral stories, and we use it as published without altering any item (Appendix~\ref{app:dataquality}). For its 797 gold-Neutral items, \texttt{normad\_v1} and \texttt{normad\_v2} disagree on the Country field in every case (they agree on every Yes/No item): v1 holds the story's source country, v2 the incidental country used to build the Neutral story, which is what the model is shown. So the Country prompt never names the country a Neutral story was written for. Cluster-level analyses use the source country to stay comparable to the benchmark's tables.

We evaluate seven models under greedy decoding, one variant per model: five open-weight and two frontier models, all listed with their reports in Table~\ref{tab:models}. Two of the families were also run with a written reasoning field, giving a paired contrast on the same weights---\texttt{gpt-5.6-luna} (frontier) and Gemma-4B (open-weight); the main analysis uses the no-reasoning variant of each and the paired contrasts are analysed separately (Section~\ref{sec:reasoning}). \texttt{gemini-3.8-flash} has hidden reasoning that cannot be turned off and also returns a post-hoc explanation, so its one variant is used as it is. The two frontier models are markedly stronger, so pooled means average two deliberately different groups; we report the two groups separately wherever the distinction matters (Table~\ref{tab:conflation}, Figure~\ref{fig:conflation}).

\begin{table*}[ht]
\centering
\small
\begin{tabular}{@{}lllllc@{}}
\toprule
Model & Shorthand & Size / type & Lab & Report & Weights \\
\midrule
Mistral-7B-Instruct-v0.2 & Mistral-7B & 7B & Mistral AI & \citep{jiang2023mistral} & \href{https://huggingface.co/mistralai/Mistral-7B-Instruct-v0.2}{link} \\
Ministral-3-3B-Instruct-2512 & Ministral-3B & 3B & Mistral AI & \citep{liu2026ministral3} & \href{https://huggingface.co/mistralai/Ministral-3-3B-Instruct-2512}{link} \\
Qwen3.5-9B & Qwen3.5-9B & 9B & Alibaba & \citep{qwen2026qwen35} & \href{https://huggingface.co/Qwen/Qwen3.5-9B}{link} \\
Qwen3.5-4B & Qwen3.5-4B & 4B & Alibaba & \citep{qwen2026qwen35} & \href{https://huggingface.co/Qwen/Qwen3.5-4B}{link} \\
\texttt{gemma-4-e4b-it} & Gemma-4B & 8B & Google & \citep{gemmateam2026gemma4} & \href{https://huggingface.co/google/gemma-4-e4b-it}{link} \\
\texttt{gpt-5.6-luna} & GPT-5.6 & Frontier & OpenAI & \citep{openai2026gpt56} & --- \\
\texttt{gemini-3.8-flash} & Gemini-3.8 & Frontier & Google DeepMind & \citep{googledeepmind2026gemini38flash} & --- \\
\bottomrule
\end{tabular}
\caption{Evaluated models and their technical reports. Weights are linked for open-weight models; the two frontier models are API-only (---).}
\label{tab:models}
\end{table*}

\subsection{Experimental design}

Table~\ref{tab:instruments} defines the seven instruments; each is run at all three context levels, so the design has 21 fully specified prompts. The original instrument is \ins{e1}. Three variants add a definition of Neither: \ins{correcteddef} defines it in one clause, \ins{full} defines all three options, and \ins{vague} gives a content-free clause that permits answering C without defining anything, which separates the effect of a definition from the effect of permission. The four-cell \ins{joint4} makes relevance an explicit part of every option. Two dedicated probes score one decision each: \ins{apply3strip\_verbose} for applicability, which states its options as full sentences and removes the story's own trailing acceptability question so that the probe does not contradict its own instruction to judge relevance, and \ins{binaryonly} for conformity, a forced Yes/No choice scored on gold-applicable items only (``gold-routed'').

\begin{table*}[ht]
\centering
\small
\setlength{\tabcolsep}{5pt}
\begin{tabular}{@{}>{\raggedright\arraybackslash}p{2.7cm}>{\raggedright\arraybackslash}p{4.4cm}>{\raggedright\arraybackslash}p{3.0cm}>{\raggedright\arraybackslash}p{2.6cm}@{}}
\toprule
Instrument & What it asks & Answer options & Items scored \\
\midrule
\ins{e1} & original acceptability question & Yes / No / Neither & all 2,615 \\
\ins{correcteddef} & \ins{e1} + one clause defining Neither & Yes / No / Neither & all 2,615 \\
\ins{full} & \ins{e1} + a clause defining all three options & Yes / No / Neither & all 2,615 \\
\ins{vague} & \ins{e1} + a content-free clause (control) & Yes / No / Neither & all 2,615 \\
\ins{joint4} & relevance and acceptability in one four-cell question & four sentences (A--D) & all 2,615 \\
\ins{apply3strip\_\allowbreak verbose} & applicability only; story's acceptability question removed & two full sentences & all 2,615 \\
\ins{binaryonly} & conformity only, forced choice & Yes / No & gold-applicable 1,818 \\
\bottomrule
\end{tabular}
\caption{The seven instruments.}
\label{tab:instruments}
\end{table*}

We use these instruments in five steps: (1) replicate the benchmark with \ins{e1}; (2) test whether defining Neither helps (\ins{correcteddef}, \ins{full}, \ins{vague}); (3) score applicability and conformity with the two probes and compare them; (4) test whether the applicability probe is independent of acceptability by comparing its NotApply rate on gold-Yes and gold-No items, which are equally applicable by construction; and (5) ask whether any single-prompt variant---\ins{joint4} or a ternary rewording---can replace the two probes. As secondary descriptive material, carrying no headline claim, the appendix also breaks accuracy down by culture cluster and subaxis and gives the Latin-America gift case, alongside worked example items (Appendices~\ref{app:tables}, \ref{app:extra}, and~\ref{app:examples}).

\subsection{Metrics and statistics}

For ternary answers we report accuracy and class-conditional recall. For the two probes we use Youden's $J$ (sensitivity $+$ specificity $-1$; zero is chance, negative values reported as they are), with NotApply as the positive class for applicability and Yes versus No on gold-applicable items for conformity. We report the two capabilities separately and never as a composite, because they come from different instruments and a composite would hide which axis failed. \emph{Conflation} is the signed gap, in percentage points, between the NotApply rate on gold-No items and that on gold-Yes items---zero when the applicability probe is independent of the acceptability label---tested with an exact test and Benjamini--Hochberg correction within each instrument's model-by-level cells. This is distinct from the ternary answer's bundling of the two decisions, where the stronger conformity masks the weaker applicability (Section~\ref{sec:bundle}). Intervals use 20,000 bootstrap resamples with seeds from a stable checksum; cluster-level accuracy uses label-balanced, country-clustered resampling.

\section{Results}

\subsection{The Country deficit is dominated by applicability}
\label{sec:country}

\begin{figure*}[ht]
\centering
\includegraphics[width=\textwidth]{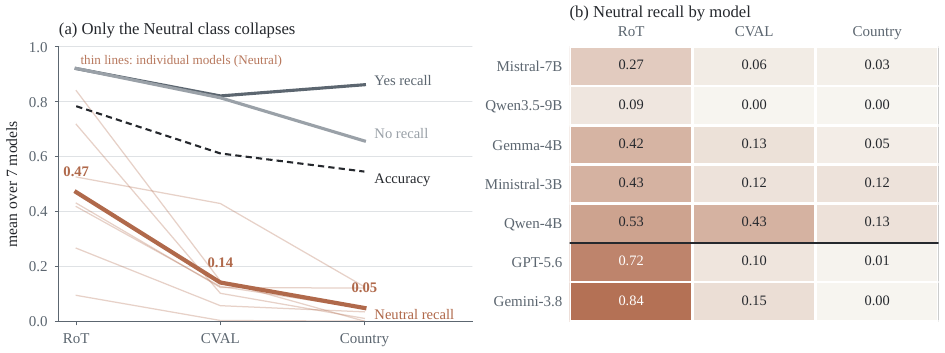}
\caption{(a) Mean class recall and accuracy under \ins{e1} as context is removed; only the Neutral line collapses (thin lines are individual models). (b) Neutral recall for each model; the Country column is boxed.}
\label{fig:ladder}
\end{figure*}

The original instrument reproduces the reported pattern for every model: Country accuracy and Neutral recall are far below RoT (accuracy in Table~\ref{tab:e1}, Neutral recall in Figure~\ref{fig:ladder}). Mean accuracy falls steadily (0.784, 0.612, 0.546), and the decline is dominated by the Neutral class: mean Neutral recall collapses from 0.470 to 0.141 to 0.048, whereas mean Yes recall stays high (0.92 to 0.86) and mean No recall falls only to 0.66. Qwen3.5-9B and Gemini-3.8 recover no gold-Neutral item at Country.

Two facts make this an applicability problem rather than a general difficulty. First, the label: the country shown for Neutral items is incidental (Section~\ref{sec:design}), so it carries no information about the label. Second, the decision: the loss is in the Neutral class while Yes/No calls stay confident, so a low Country accuracy is mostly a \emph{does not apply} judgment read as an acceptability verdict. Appendix~\ref{app:examples} works through a gold-Neutral item, giving the rule, the incidental country, and the models' answers at each level.

\begin{table}[ht]
\centering
\small
\begin{tabular}{@{}lrrr@{}}
\toprule
Model & RoT acc. & CVAL acc. & Country acc. \\
\midrule
Mistral-7B & .745 & .619 & .557 \\
Qwen3.5-9B & .706 & .622 & .564 \\
Gemini-3.8 & .928 & .715 & .653 \\
Gemma-4B & .802 & .627 & .495 \\
GPT-5.6 & .895 & .626 & .559 \\
Ministral-3B & .659 & .522 & .435 \\
Qwen3.5-4B & .754 & .550 & .555 \\
\midrule
\textbf{mean} & 0.784 & 0.612 & 0.546 \\
\bottomrule
\end{tabular}
\caption{Replication under \ins{e1}, the benchmark's original Yes/No/Neither question (the exact prompt is given in Appendix~\ref{app:base}): accuracy at each context level. Neutral recall is in Figure~\ref{fig:ladder}.}
\label{tab:e1}
\end{table}

\begin{figure*}[ht]
\centering
\includegraphics[width=\textwidth]{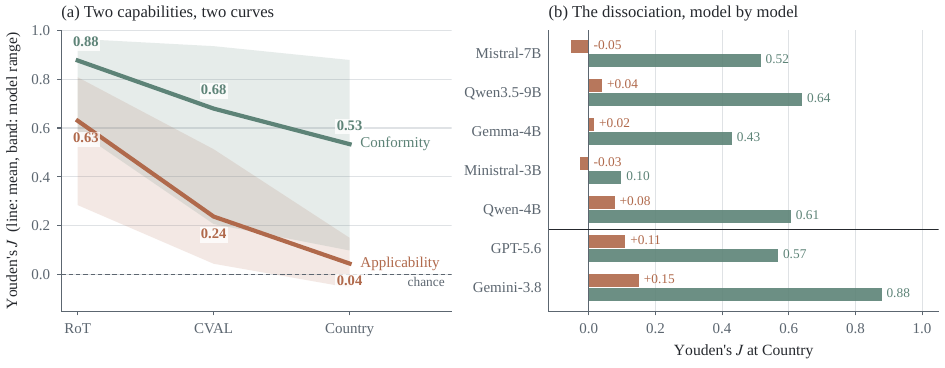}
\caption{(a) Mean Youden's $J$ for applicability and conformity by context; bands span the seven models. Applicability reaches chance at Country, conformity does not. (b) The same dissociation model by model at Country.}
\label{fig:capabilities}
\end{figure*}

The dedicated probes confirm that this is specifically applicability. Figure~\ref{fig:capabilities} and Table~\ref{tab:capconf} put the two capabilities on the same items and models. Applicability is detectable in 18 of 21 cells and falls steeply with context (mean $J$ 0.629, 0.237, 0.045); the three failures are all at Country, where Mistral-7B ($-0.054$) and Ministral-3B ($-0.027$) are significantly \emph{below} chance and Gemma-4B (0.016) is indistinguishable from it. Conformity is above chance in all 21 cells and falls only from 0.876 to 0.534. From RoT to Country, applicability loses about 0.58 $J$ and conformity about 0.34, so a country name supports conformity far better than applicability. The same pattern holds per model (Table~\ref{tab:capconf}): every model's applicability drops to near chance at Country, while its conformity stays well above it.

\paragraph{The weakest model.} Ministral-3B is the smallest and weakest model in the roster and the most degenerate: under forced choice it rejects almost every item (Country false-reject rate 0.90; Table~\ref{tab:conffull}), so its conformity $J$ is the lowest and it is one of the two models that fall below chance on Country applicability. This is a property of that one model, not a driver of the audit: the Country applicability collapse, the conflation gap, and conformity above chance at every level all hold across the other open models, and no headline pattern depends on Ministral-3B.

\begin{table*}[ht]
\centering
\small
\begin{tabular}{@{}lrrrrrr@{}}
\toprule
& \multicolumn{3}{c}{Applicability $J$} & \multicolumn{3}{c}{Conformity $J$} \\
\cmidrule(lr){2-4}\cmidrule(lr){5-7}
Model & RoT & CVAL & Country & RoT & CVAL & Country \\
\midrule
Mistral-7B & +.485 & +.160 & $-.054^\dagger$ & .851 & .688 & .516 \\
Qwen3.5-9B & +.755 & +.108 & +.040 & .916 & .772 & .640 \\
Gemini-3.8 & +.807 & +.513 & +.150 & .965 & .935 & .878 \\
Gemma-4B & +.546 & +.292 & +.016 & .941 & .728 & .430 \\
GPT-5.6 & +.808 & +.381 & +.110 & .964 & .719 & .568 \\
Ministral-3B & +.284 & +.044 & $-.027^\dagger$ & .587 & .207 & .098 \\
Qwen3.5-4B & +.721 & +.163 & +.078 & .910 & .707 & .606 \\
\bottomrule
\end{tabular}
\caption{Per-model applicability $J$ (\ins{apply3strip\_verbose}) and conformity $J$ (\ins{binaryonly}, gold-routed); $^\dagger$ marks Country cells whose interval lies below zero.}
\label{tab:capconf}
\end{table*}

\subsection{The ternary answer bundles the two decisions}
\label{sec:bundle}

The Country result is visible only because the two decisions are scored separately. In the benchmark's single answer they are not: because conformity is the stronger signal, the score follows it and hides the weaker applicability, which we call \emph{bundling}. Two pieces of evidence show it. First, the aggregate is carried by conformity: the Yes/No rows stay strong at every level (mean $J$ 0.83 and 0.74 at RoT, still 0.46 and 0.43 at Country) while the Neutral row is already weak at RoT ($+0.42$) and zero at Country ($-0.01$); a model that answers ``Yes'' or ``No'' on an item whose gold label is ``does not apply'' is not making a conformity error, it never took the applicability step. Second, the two capabilities are measured with different instruments and are demonstrably dependent in the small open models (Section~\ref{sec:conflation}); a single number therefore cannot be read as either one.

\subsection{Defining Neither helps where the rule is available}
\label{sec:wording}

\begin{figure*}[ht]
\centering
\includegraphics[width=\textwidth]{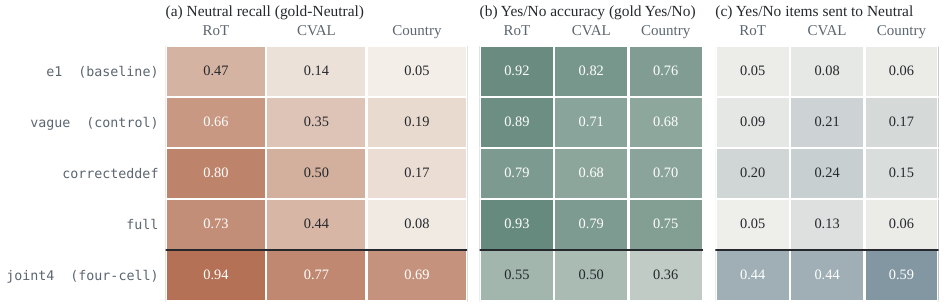}
\caption{Means over seven models for each prompt (rows) and context level (columns). (a) Neutral recall rises at RoT and CVAL for the clause variants but barely at Country; \ins{vague} is a content-free control. (b) The cost in Yes/No accuracy. (c) The share of gold Yes/No items answered Neutral: \ins{joint4} reaches high Neutral recall by sending most applicable items there.}
\label{fig:wording}
\end{figure*}

Defining Neither raises mean Neutral recall by 0.333 at RoT and 0.358 at CVAL, and the frontier models reach about 0.87 at RoT once it is defined (Table~\ref{tab:wording}, Figure~\ref{fig:wording}). The gain is larger at RoT than at Country for six of seven models. At Country it is only 0.119, which is what underdetermination predicts: there is no rule in the prompt for the definition to attach to. A content-free control that defines nothing still reaches 0.66 Neutral recall at RoT, so part of the RoT and CVAL gain is permission to answer Neither rather than a sharper definition, and at Country the full definition even yields less Neutral recall than the control for four of seven models. The improvement is not free: mean Yes/No accuracy falls by 0.130, 0.134, and 0.061.

\begin{table}[ht]
\centering
\small
\begin{tabular}{@{}lrr@{}}
\toprule
Level & Neutral recall gain & Yes/No acc.\ change \\
\midrule
RoT & $+0.333$ & $-0.130$ \\
CVAL & $+0.358$ & $-0.134$ \\
COUNTRY & $+0.119$ & $-0.061$ \\
\bottomrule
\end{tabular}
\caption{Mean change from \ins{e1} to \ins{correcteddef} over seven models.}
\label{tab:wording}
\end{table}

A different format fares no better as a single replacement. Across \ins{e1}, \ins{correcteddef}, \ins{vague}, \ins{full}, and the four-cell \ins{joint4}, the mean Country Neutral $J$ stays between $-0.01$ and $+0.10$ (Table~\ref{tab:singleprompt}). The four-cell prompt, which makes relevance explicit in every option, raises Country Neutral recall to 0.69---but only by also sending 59\% of gold Yes/No items to Neutral, with Yes/No accuracy falling to 0.36. A Neutral-recall gain bought by routing most applicable items to Neutral is not better applicability, and it is why we use no ternary variant as the measurement of record.

\begin{table*}[ht]
\centering
\small
\begin{tabular}{@{}lrrrr@{}}
\toprule
Approach & \textbf{Country notapply $J$} & Neutral recall & Yes/No accuracy & Yes/No $\to$ Neutral \\
\midrule
\ins{e1} & $-0.01$ & .05 & .76 & .06 \\
\ins{correcteddef} & $+0.01$ & .17 & .70 & .15 \\
\ins{full} & $+0.02$ & .08 & .75 & .06 \\
\ins{vague} & $+0.02$ & .19 & .68 & .17 \\
\ins{joint4} & $+0.10$ & .69 & .36 & .59 \\
\bottomrule
\end{tabular}
\caption{Country-level summary of each retained ternary approach over seven models. The last column is the share of gold Yes/No items answered Neutral.}
\label{tab:singleprompt}
\end{table*}

\subsection{Clean separation at the frontier, conflation in the small models}
\label{sec:conflation}

\begin{figure*}[ht]
\centering
\includegraphics[width=\textwidth]{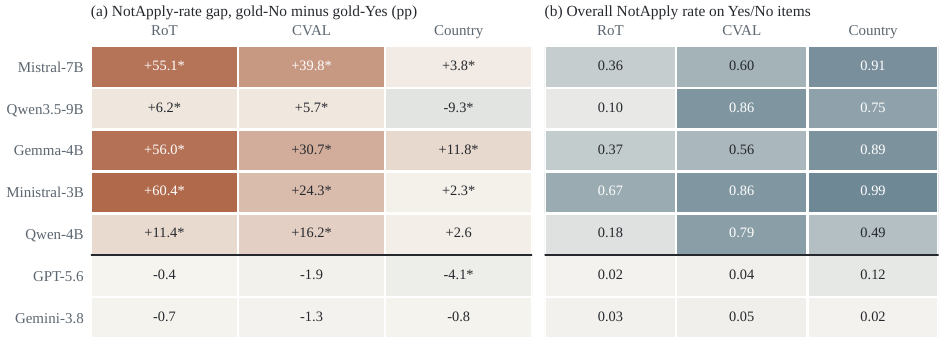}
\caption{(a) Gap between the NotApply rate on gold-No and gold-Yes items, in percentage points; $*$ marks cells significant after BH correction. Under a valid applicability probe the gap is zero. (b) Each model's overall NotApply rate on Yes/No items, across the seven models.}
\label{fig:conflation}
\end{figure*}

The two-probe design separates the two decisions and is what makes the Country asymmetry visible. For the applicability probe to measure relevance, its answer must not depend on acceptability: gold-Yes and gold-No items are equally applicable, so their NotApply rates should match. The separation is clean at the frontier level and breaks down in the small models. \texttt{gemini-3.8-flash} (frontier, reasons internally) shows no significant conflation at any level, and \texttt{gpt-5.6-luna} (frontier, does not reason) shows none at RoT or CVAL, where the applicability question is well-posed; Luna's single significant cell is a small \emph{negative} Country gap ($-4.1$\,pp), a level where the prompt supplies no rule for a Neutral label to attach to and applicability is barely measurable, so we do not read that cell as conflation. The five small open models, none of which reasons, are the opposite: their probe rate differs significantly in 14 of 15 model-by-level cells after correction (Table~\ref{tab:conflation}, Figure~\ref{fig:conflation}), almost all in the conflating direction (the lone exception is Qwen3.5-9B at Country, a small negative gap read with the saturation caveat below). Gaps are largest at RoT (mean $+37.8$\,pp over the five open models) and shrink at CVAL ($+23.3$\,pp). At Country the open models' NotApply rates on applicable items saturate (0.49--0.99) while the frontier models' collapse (0.02--0.12), so a small gap there reflects lack of variation rather than a clean probe.

\begin{table*}[ht]
\centering
\small
\begin{tabular}{@{}lrrr@{}}
\toprule
Model & RoT & CVAL & Country \\
\midrule
Mistral-7B & $+55.1^{*}$ & $+39.8^{*}$ & $+3.8^{*}$ \\
Qwen3.5-9B & $+6.2^{*}$ & $+5.7^{*}$ & $-9.3^{*}$ \\
Gemini-3.8 & $-0.7$ & $-1.3$ & $-0.8$ \\
Gemma-4B & $+56.0^{*}$ & $+30.7^{*}$ & $+11.8^{*}$ \\
GPT-5.6 & $-0.4$ & $-1.9$ & $-4.1^{*}$ \\
Ministral-3B & $+60.4^{*}$ & $+24.3^{*}$ & $+2.3^{*}$ \\
Qwen3.5-4B & $+11.4^{*}$ & $+16.2^{*}$ & $+2.6$ \\
\midrule
open models (mean) & $+37.8$ & $+23.3$ & $+2.2$ \\
frontier models (mean) & $-0.6$ & $-1.6$ & $-2.5$ \\
\bottomrule
\end{tabular}
\caption{Conflation gap in percentage points (NotApply rate on gold-No minus gold-Yes); $^{*}$ = BH-significant.}
\label{tab:conflation}
\end{table*}

We read this conflation as consistent with a representational limit in the small models, not a property of the task or a flaw in the probes. It is not specific to this instrument: it moves sharply across wordings. Within the five small models the gap is not significantly correlated with size or competence ($r=-0.45$, $p=0.44$, $n=5$), but that test is underpowered and its sign is consistent with smaller models conflating more; the reading does not rest on a within-roster size trend but on the coarse frontier-versus-small difference and on the rationale contrast. Nor is it a consequence of not reasoning alone: \texttt{gpt-5.6-luna}, a frontier model that does not reason, is clean. The rationale runs sharpen the asymmetry (Section~\ref{sec:reasoning}): the frontier model without a rationale is already clean, whereas adding a rationale to a small model moves it toward separation without removing the conflation. With only two frontier models, and with size, training, and reasoning covarying across the small models, we state this as suggestive, not demonstrated. What the data do establish is that the two-probe design separates the capabilities at the frontier level, while in the small open models the applicability probe is not independent of the label; we therefore report the conflation gap, signed, beside every applicability score rather than calling it a leak.

\subsection{The written-rationale effect in one frontier and one open-weight model}
\label{sec:reasoning}

We ask two models to write a short rationale before the answer, without enabling the provider's hidden chain-of-thought, and compare each against the same model answering directly: one frontier (\texttt{gpt-5.6-luna}) and one open-weight (Gemma-4B). Each pair uses the same weights, and neither pair enters a number above. Full tables are in Appendix~\ref{app:reasoning}.

For \texttt{gpt-5.6-luna} we only ask for a reasoning field before the answer, and the apply probe stays clean: its conflation gap remains near zero, at most $+2.2$\,pp, against $+6$ to $+60$\,pp for the small open models.

For Gemma-4B we use the same weights but ask for a reasoning field as well (the two runs differ slightly in serving stack: a single-token GBNF completion from a BF16 GGUF versus transformers with outlines JSON). The one change that matters here is that its RoT conflation gap drops sharply, though it does not vanish: from $+56.0$ to $+15.2$\,pp, a change of $-40.8$\,pp.

So a written rationale roughly quarters the small model's conflation without removing it. Other numbers change only modestly and are not load-bearing for the audit, so they are relegated to Appendix~\ref{app:reasoning}.

\section{Discussion}

The Country deficit is dominated by applicability: the prompt withholds the relation that makes a Neutral item Neutral, so applicability falls to chance there while conformity falls far less (Section~\ref{sec:country}). The Country score is therefore evidence not only that models lack cultural knowledge but also that the benchmark asks a less answerable question when the rule is withheld.

The second claim is about that format: one answer stands in for both decisions, so the score is carried by the easier conformity (Section~\ref{sec:bundle}), and scoring each decision on its own separates them at the frontier level but leaves conflation in the small open models (Section~\ref{sec:conflation}) that an applicability score must be reported alongside. A successor benchmark should publish an explicit applicability label, define each answer cell, report conformity only under a declared applicability rule, and report the two results separately with the conflation gap.

\section{Conclusion}

NormAd-ETI's Neutral result is not one phenomenon. At Country the prompt withholds the rule that defines a Neutral label, and applicability is what falls furthest: measured with separate probes it reaches chance while conformity stays well above it. The single ternary answer bundles the two decisions, so the score is carried by conformity; scoring each on its own separates them at the frontier level, while the small open models retain conflation consistent with a representational limit in the small models. We recommend treating the ternary accuracy as a legacy diagnostic and reporting the two results separately with the conflation gap. More generally, cultural benchmarks should validate not only whether labels are reproducible, but also whether the label names, construction procedure, prompts, and reported interpretations refer to the same construct.

\section*{Limitations}

\paragraph{Scope.} This is an analysis of a single benchmark, using our own model runs across the instruments of Section~\ref{sec:design}. Its main limitation is generality: the findings are about NormAd-ETI and need not transfer to other cultural-reasoning datasets, and we do not reannotate the data with independent human judgments.

\paragraph{No human baseline.} The Country condition may contain genuine cultural-knowledge demands in addition to the applicability problem, and the benchmark reports no human baseline for Country alone. We therefore claim underdetermination by the prompt, not a human gap.

\paragraph{Measurement, not mechanism.} The applicability/conformity decomposition is a measurement design, not a claim about processes inside a model; the reading of the probe's conflation as a representational limit in the small models is suggestive, not demonstrated.

\paragraph{Model set and the separation gap.} The set mixes five small open models with two strong frontier models, so pooled means are driven by between-group differences. The two frontier models differ from the open models in size and capability, and only \texttt{gemini-3.8-flash} reasons internally while \texttt{gpt-5.6-luna} does not; both frontier models are nonetheless clean, so the absence of reasoning alone does not create conflation, and we cannot fully separate size, training, and reasoning as causes.

\paragraph{Written rationale.} The rationale runs are analysed separately and enter no main number. Two paired contrasts use the same weights within each pair---\texttt{gpt-5.6-luna} (frontier) and Gemma-4B (open-weight). Luna's two runs share the API stack and differ only by the reasoning field; Gemma's no-reasoning run is a single-token GBNF completion from a BF16 GGUF while its reasoning run uses transformers with outlines JSON decoding, so that pair also varies in serving stack. There is one run per condition, so the paired differences include any nondeterminism at temperature 0, and \texttt{gemini-3.8-flash} has no no-rationale counterpart.

\paragraph{Cluster mapping.} The Inglehart--Welzel mapping is used as published; it places Malta and the Philippines in Latin America, and removing them shifts the cluster's balanced accuracy by 0.9\,pp and its gift false-accept rate by 2.2\,pp on 16--22 items (Appendix~\ref{app:extra}).

\paragraph{Data-quality heuristics.} The story-leak screening is a marker heuristic that flags 10 of the 219 Neutral stories the authors regenerated in the corrected release (against 1 of the same 219 before regeneration); removing them changes no cluster mean by more than 0.27\,pp (Appendix~\ref{app:extra}).

\bibliography{references}

\appendix
\section{Full result tables}
\label{app:tables}

This appendix reports every per-model statistic behind the paper. Each subsection states what its table contains and how the columns are computed. The exact prompts are given in Appendix~\ref{app:prompts}.

\subsection{The \ins{e1} baseline (Table~\ref{tab:e1full})}

Table~\ref{tab:e1full} expands the \ins{e1} baseline: overall accuracy and recall for the Yes, No, and Neutral classes, for each of the seven models at each context level. Accuracy falls as context is removed and the fall is dominated by the Neutral class, on which two models score zero recall at Country.

\begin{table*}[ht]
\centering
\small
\begin{tabular}{llrrrr}
\toprule
Model & Level & Acc. & R(Yes) & R(No) & R(Neutral) \\
\midrule
Mistral-7B & RoT & .745 & .980 & .927 & .266 \\
 & CVAL & .619 & .938 & .787 & .056 \\
 & COUNTRY & .557 & .938 & .622 & .034 \\
Qwen3.5-9B & RoT & .706 & .959 & .990 & .094 \\
 & CVAL & .622 & .900 & .887 & .003 \\
 & COUNTRY & .564 & .927 & .688 & .000 \\
Gemini-3.8 & RoT & .928 & .963 & .969 & .841 \\
 & CVAL & .715 & .969 & .960 & .147 \\
 & COUNTRY & .653 & .962 & .914 & .000 \\
Gemma-4B & RoT & .802 & .973 & .967 & .418 \\
 & CVAL & .627 & .845 & .848 & .127 \\
 & COUNTRY & .495 & .940 & .425 & .046 \\
GPT-5.6 & RoT & .895 & .983 & .962 & .718 \\
 & CVAL & .626 & .965 & .739 & .102 \\
 & COUNTRY & .559 & .965 & .623 & .010 \\
Ministral-3B & RoT & .659 & .846 & .665 & .430 \\
 & CVAL & .522 & .718 & .675 & .123 \\
 & COUNTRY & .435 & .524 & .626 & .120 \\
Qwen3.5-4B & RoT & .754 & .744 & .971 & .526 \\
 & CVAL & .550 & .414 & .806 & .429 \\
 & COUNTRY & .555 & .780 & .703 & .125 \\
\bottomrule
\end{tabular}
\caption{\ins{e1} accuracy and per-class recall.}
\label{tab:e1full}
\end{table*}

\subsection{Applicability (Table~\ref{tab:appfull})}

Applicability $J$ from \ins{apply3strip\_verbose}, with a 20{,}000-resample bootstrap interval, the signed conflation gap in percentage points, its log odds ratio, and whether the exact test survives Benjamini--Hochberg correction within the seven model $\times$ three level cells. $J$ is unfloored, so the two negative Country cells are reported as estimated rather than clamped to zero.

\begin{table*}[ht]
\centering
\small
\begin{tabular}{llrrrrr}
\toprule
Model & Level & $J$ & 95\% CI & gap (pp) & log OR & sig. \\
\midrule
Mistral-7B & RoT & +0.485 & [.451,.519] & +55.1 & 2.85 & * \\
 & CVAL & +0.160 & [.122,.197] & +39.8 & 1.79 & * \\
 & COUNTRY & $-$0.054 & [$-$0.083,$-$0.027] & +3.8 & 0.47 & * \\
Qwen3.5-9B & RoT & +0.755 & [.727,.783] & +6.2 & 0.68 & * \\
 & CVAL & +0.108 & [.088,.128] & +5.7 & 0.49 & * \\
 & COUNTRY & +0.040 & [.006,.075] & $-$9.3 & $-$0.49 & * \\
Gemini-3.8 & RoT & +0.807 & [.779,.834] & $-$0.7 & $-$0.24 & \\
 & CVAL & +0.513 & [.478,.549] & $-$1.3 & $-$0.28 & \\
 & COUNTRY & +0.150 & [.124,.178] & $-$0.8 & $-$0.39 & \\
Gemma-4B & RoT & +0.546 & [.517,.575] & +56.0 & 2.82 & * \\
 & CVAL & +0.292 & [.259,.326] & +30.7 & 1.29 & * \\
 & COUNTRY & +0.016 & [$-$0.009,.042] & +11.8 & 1.35 & * \\
GPT-5.6 & RoT & +0.808 & [.780,.835] & $-$0.4 & $-$0.19 & \\
 & CVAL & +0.381 & [.345,.416] & $-$1.9 & $-$0.51 & \\
 & COUNTRY & +0.110 & [.077,.144] & $-$4.1 & $-$0.39 & * \\
Ministral-3B & RoT & +0.284 & [.258,.310] & +60.4 & 4.51 & * \\
 & CVAL & +0.044 & [.017,.069] & +24.3 & 3.03 & * \\
 & COUNTRY & $-$0.027 & [$-$0.042,$-$0.012] & +2.3 & 2.70 & * \\
Qwen3.5-4B & RoT & +0.721 & [.693,.748] & +11.4 & 0.78 & * \\
 & CVAL & +0.163 & [.138,.187] & +16.2 & 1.01 & * \\
 & COUNTRY & +0.078 & [.035,.119] & +2.6 & 0.10 & \\
\bottomrule
\end{tabular}
\caption{Applicability from \ins{apply3strip\_verbose}: $J$, interval, conflation gap in percentage points, its log odds ratio, and BH significance ($*$).}
\label{tab:appfull}
\end{table*}

\subsection{Conformity (Table~\ref{tab:conffull})}

Conformity $J$ from \ins{binaryonly}, scored only on the 1,818 gold-applicable items so that the model's own applicability call cannot bias the sample. False accept is the rate of accepting gold No items and false reject the rate of rejecting gold Yes items; separating them shows that similar $J$ values can come from opposite failures (Ministral-3B rejects nearly everything; Gemma-4B accepts violations).

\begin{table*}[ht]
\centering
\small
\begin{tabular}{llrrrr}
\toprule
Model & Level & $J$ & 95\% CI & False accept & False reject \\
\midrule
Mistral-7B & RoT & .851 & [.827,.875] & .139 & .010 \\
 & CVAL & .688 & [.656,.719] & .271 & .041 \\
 & COUNTRY & .516 & [.481,.551] & .449 & .035 \\
Qwen3.5-9B & RoT & .916 & [.898,.933] & .003 & .081 \\
 & CVAL & .772 & [.743,.800] & .074 & .154 \\
 & COUNTRY & .640 & [.606,.676] & .270 & .090 \\
Gemini-3.8 & RoT & .965 & [.952,.977] & .015 & .020 \\
 & CVAL & .935 & [.919,.951] & .032 & .033 \\
 & COUNTRY & .878 & [.856,.900] & .082 & .039 \\
Gemma-4B & RoT & .941 & [.925,.956] & .016 & .043 \\
 & CVAL & .728 & [.697,.759] & .104 & .168 \\
 & COUNTRY & .430 & [.393,.466] & .512 & .058 \\
GPT-5.6 & RoT & .964 & [.951,.976] & .026 & .010 \\
 & CVAL & .719 & [.688,.749] & .253 & .029 \\
 & COUNTRY & .568 & [.533,.602] & .402 & .030 \\
Ministral-3B & RoT & .587 & [.555,.619] & .000 & .413 \\
 & CVAL & .207 & [.180,.233] & .002 & .791 \\
 & COUNTRY & .098 & [.079,.118] & .003 & .898 \\
Qwen3.5-4B & RoT & .910 & [.891,.928] & .007 & .083 \\
 & CVAL & .707 & [.676,.737] & .059 & .233 \\
 & COUNTRY & .606 & [.570,.642] & .193 & .200 \\
\bottomrule
\end{tabular}
\caption{Conformity from \ins{binaryonly} on gold-applicable items.}
\label{tab:conffull}
\end{table*}

\subsection{Cross-cultural accuracies (Table~\ref{tab:clusters})}

Label-balanced accuracy by Inglehart--Welzel cluster, resampled by country rather than by item, with 95\% intervals; $n$ is the number of countries in the cluster. This is a descriptive breakdown only, not used to support a headline claim: the Confucian-versus-Latin-America comparison is the only near-separation and should be read as marginal.

\begin{table*}[ht]
\centering
\small
\begin{tabular}{lrrrr}
\toprule
Cluster & $n$ countries & Country acc.\ [95\% CI] & CVAL acc.\ [95\% CI] \\
\midrule
Confucian & 5 & 0.580 [.533,.625] & 0.630 [.583,.677] \\
Protestant Europe & 4 & 0.560 [.513,.596] & 0.589 [.548,.622] \\
English Speaking & 6 & 0.551 [.515,.583] & 0.628 [.602,.656] \\
Catholic Europe & 7 & 0.539 [.496,.580] & 0.606 [.575,.636] \\
West and South Asia & 18 & 0.526 [.498,.552] & 0.606 [.580,.631] \\
Orthodox Europe & 7 & 0.504 [.468,.541] & 0.572 [.549,.594] \\
African-Islamic & 19 & 0.504 [.472,.535] & 0.578 [.551,.603] \\
Latin America & 9 & 0.492 [.451,.531] & 0.557 [.524,.588] \\
\bottomrule
\end{tabular}
\caption{Label-balanced accuracy by Inglehart--Welzel cluster with 95\% country-clustered intervals.}
\label{tab:clusters}
\end{table*}

\subsection{Where the benchmark is hardest}
\label{sec:culture}

Pooling over the seven models with label-balanced, country-clustered intervals, gifts is the hardest subaxis at every level (Table~\ref{tab:subaxis}). These breakdowns show that an aggregate hides model-specific failure directions; they do not by themselves test the underdetermination.

\begin{table}[ht]
\centering
\small
\begin{tabular}{@{}lrrr@{}}
\toprule
Subaxis & RoT & CVAL & Country \\
\midrule
gifts & 0.739 & 0.541 & 0.472 \\
visiting & 0.760 & 0.612 & 0.538 \\
eating & 0.785 & 0.599 & 0.537 \\
basic\_etiquette & 0.793 & 0.613 & 0.539 \\
\bottomrule
\end{tabular}
\caption{Label-balanced accuracy by subaxis, pooled over seven models.}
\label{tab:subaxis}
\end{table}

\subsection{Ternary variants (Tables~\ref{tab:ternaryj} and~\ref{tab:ternaryacc})}

Table~\ref{tab:ternaryj} gives mean per-class Youden's $J$ for every retained ternary approach, over all seven models, treating Neutral as ``notapply'', Yes as ``conform'', and No as ``noconform''. Table~\ref{tab:ternaryacc} gives the cost side; the last column, the share of gold Yes/No items answered Neutral, rises from 0.05--0.08 for \ins{e1} to 0.44--0.60 for \ins{joint4}.

\begin{table*}[ht]
\centering
\footnotesize
\setlength{\tabcolsep}{3pt}
\begin{tabular}{@{}llccc@{}}
\toprule
Approach & Class & RoT & CVAL & Country \\
\midrule
\ins{e1} & notapply & $+0.42$ & $+0.06$ & $-0.01$ \\
 & conform & $+0.83$ & $+0.57$ & $+0.46$ \\
 & noconform & $+0.74$ & $+0.56$ & $+0.43$ \\
\ins{correcteddef} & notapply & $+0.61$ & $+0.26$ & $+0.01$ \\
 & conform & $+0.77$ & $+0.56$ & $+0.46$ \\
 & noconform & $+0.70$ & $+0.49$ & $+0.39$ \\
\ins{full} & notapply & $+0.68$ & $+0.31$ & $+0.02$ \\
 & conform & $+0.87$ & $+0.65$ & $+0.46$ \\
 & noconform & $+0.84$ & $+0.57$ & $+0.40$ \\
\ins{joint4} & notapply & $+0.49$ & $+0.33$ & $+0.10$ \\
 & conform & $+0.62$ & $+0.46$ & $+0.30$ \\
 & noconform & $+0.43$ & $+0.37$ & $+0.22$ \\
\ins{vague} & notapply & $+0.57$ & $+0.13$ & $+0.02$ \\
 & conform & $+0.85$ & $+0.55$ & $+0.48$ \\
 & noconform & $+0.76$ & $+0.46$ & $+0.34$ \\
\bottomrule
\end{tabular}
\caption{Mean per-class Youden's $J$ for each ternary approach (seven models; $\pm$0.01--0.02).}
\label{tab:ternaryj}
\end{table*}

\begin{table*}[ht]
\centering
\footnotesize
\setlength{\tabcolsep}{3pt}
\begin{tabular}{@{}lccc@{}}
\toprule
Approach & Yes/No acc. & Neut. recall & Y/N $\to$ Neut. \\
\midrule
\ins{e1} & .92/.82/.76 & .47/.14/.05 & .05/.08/.06 \\
\ins{correcteddef} & .79/.68/.70 & .80/.50/.17 & .20/.24/.15 \\
\ins{full} & .93/.79/.75 & .73/.44/.08 & .05/.13/.06 \\
\ins{joint4} & .55/.50/.36 & .94/.77/.69 & .44/.44/.59 \\
\ins{vague} & .89/.71/.68 & .66/.35/.19 & .09/.21/.17 \\
\bottomrule
\end{tabular}
\caption{Ternary variants, values as RoT/CVAL/Country. ``Y/N $\to$ Neut.'' is the share of gold Yes/No items sent to Neutral.}
\label{tab:ternaryacc}
\end{table*}

\section{Additional breakdowns}
\label{app:extra}

Table~\ref{tab:latam} counts, for each model, how many of the 22 Latin-America gift-violation items it accepts at Country---gold-No violations that a model wrongly endorses when only the country is shown. The spread from 22 to 1 shows large variation between models on these items; because the items come from a single region, the table is descriptive and does not by itself compare regions.

\begin{table}[ht]
\centering
\small
\begin{tabular}{@{}lr@{}}
\toprule
Model & Gift violations accepted (of 22) \\
\midrule
Gemma-4B & 22 \\
Mistral-7B & 20 \\
GPT-5.6 & 19 \\
Qwen3.5-9B & 15 \\
Qwen3.5-4B & 11 \\
Ministral-3B & 2 \\
Gemini-3.8 & 1 \\
\bottomrule
\end{tabular}
\caption{Latin-America gift-violation false accepts at Country, per model.}
\label{tab:latam}
\end{table}

\subsection{Data-quality checks}
\label{app:dataquality}

Two sensitivity checks bound the data-quality caveats in the Limitations. First, a marker heuristic on the story text flags 10 of the 219 gold-Neutral stories regenerated in the corrected release as containing a leak of the acceptability judgment, against 1 of the same 219 before regeneration; excluding the flagged items leaves every cluster mean unchanged by more than 0.27\,pp. Second, the Inglehart--Welzel map is used as published: it places Malta and the Philippines in Latin America, and excluding those two countries moves the Latin-America label-balanced accuracy (Table~\ref{tab:clusters}) by 0.9\,pp and its gift false-accept rate by 2.2\,pp, on 16--22 items.

\section{Example items}
\label{app:examples}

Table~\ref{tab:examples} shows three annotated items from the corrected release, one of each gold label, with the context the model is shown and its \ins{e1} answer at each level. The Neutral item (ID~1818) matters most here. Its paired Rule-of-Thumb is about dressing neatly and avoiding old clothes, which has nothing to do with its story about greeting a host and thanking her before leaving, so the rule does not apply and the item is gold-Neutral. The country shown to the model (Peru) is the incidental country from the regenerated release, not the story's source country (Austria). At RoT the rule is visible and the frontier model answers Neutral correctly, but at CVAL and Country it answers Yes and No instead, and the open model never answers Neutral at all: once the relevant rule is no longer identified, the action is judged as (un)acceptable. The No item shows the same failure from the other side: the rule is withheld and a violation is accepted at Country; the Yes item stays correct throughout.

\begin{table*}[ht]
\centering
\footnotesize
\setlength{\tabcolsep}{3pt}
\begin{tabular}{@{}p{3.4cm}lp{1.7cm}p{3.4cm}p{1.8cm}p{1.8cm}@{}}
\toprule
Story (abridged) & Gold & Country shown & Rule / value shown & GPT-5.6 & Mistral-7B \\
\midrule
Maria arrives early, greets the host warmly, and thanks her before leaving. (ID~1818) & Neutral & peru (src.\ austria) & RoT: ``dress neatly, no old or dirty clothes''; Value: personal presentation & Neu.\ / Yes / No & Yes / Yes / Yes \\
Alice confirms the time but arrives ten minutes late. (ID~171) & No & U.K. & RoT: ``plan ahead and arrive within the agreed time''; Value: respect for the host's time & Yes / Yes / Yes & No / No / Yes \\
Sarah greets everyone present before sitting down. (ID~0) & Yes & egypt & RoT: ``greet everyone present before starting any interaction'' & Yes / Yes / Yes & Yes / Yes / Yes \\
\bottomrule
\end{tabular}
\caption{Three example items (one per gold label) with the \ins{e1} answer at RoT\,/\,CVAL\,/\,Country. For the Neutral item the paired rule does not apply and the shown country (Peru) is incidental, not the story's source (Austria); the rule is withheld entirely at Country. ``Neu.'' is the Neutral answer. Items are from the corrected \texttt{normad\_v2} release.}
\label{tab:examples}
\end{table*}

\section{The reasoning field}
\label{app:reasoning}

The written-rationale variants are paired with their no-rationale counterparts on identical items and enter no main-analysis number. Two same-weight pairs are compared, one frontier (\texttt{gpt-5.6-luna}) and one open-weight (Gemma-4B); both are treated as clean pairs, with the Gemma runs differing slightly in serving stack (Section~\ref{sec:reasoning}). Reading convention: every difference is (rationale run) minus (no-rationale run); $*$ marks Benjamini--Hochberg significance over the 5 headline metrics $\times$ 2 pairs $\times$ 3 levels. Table~\ref{tab:outputformat} records the output format and decoding stack of every run.

\begin{table*}[ht]
\centering
\footnotesize
\begin{tabular}{@{}>{\raggedright\arraybackslash}p{3.4cm}>{\raggedright\arraybackslash}p{5.0cm}>{\raggedright\arraybackslash}p{5.0cm}@{}}
\toprule
Run & Output & Stack \\
\midrule
GPT-5.6 & \texttt{\{"answer"\}} & API, JSON schema \\
GPT-5.6 + rationale & \texttt{\{"reasoning", "answer"\}} & API, JSON schema; rationale first \\
Gemma-4B & one answer letter & llama.cpp, GBNF, BF16 GGUF \\
Gemma-4B + rationale & \texttt{\{"reasoning", "answer"\}} & transformers + outlines JSON \\
Gemini-3.8 & hidden \texttt{\textless think\textgreater} tokens, then \texttt{\{"answer", "reasoning"\}} & API; rationale post-hoc \\
Mistral-7B, Qwen3.5, Qwen3.5-4B, Ministral-3B & one answer letter & llama.cpp, GBNF, BF16 GGUF \\
\bottomrule
\end{tabular}
\caption{Output format and decoding stack per run. The open-weight runs return a single answer letter (\ins{A}/\ins{B}/\ins{C} for the \ins{e1}-style instruments, \ins{A}/\ins{B} for \ins{binaryonly} and \ins{apply3strip\_verbose}, \ins{A}/\ins{B}/\ins{C}/\ins{D} for \ins{joint4}); the frontier runs return JSON. ``GPT-5.6 + rationale'' and ``Gemma-4B + rationale'' are the reasoning runs; Gemini-3.8's text is post-hoc and it has no no-reasoning pair.}
\label{tab:outputformat}
\end{table*}

\begin{table*}[ht]
\centering
\small
\setlength{\tabcolsep}{6pt}
\begin{tabular}{@{}lrrr@{}}
\toprule
\multicolumn{4}{@{}l}{GPT-5.6 (rationale) minus GPT-5.6 (option only)} \\
Metric & RoT & CVAL & Country \\
\midrule
e1 accuracy & $-0.025^{*}$ & $-0.001$ & $+0.003$ \\
e1 Neutral recall & $-0.090^{*}$ & $-0.019$ & $+0.015^{*}$ \\
Conformity $J$ & $-0.008$ & $+0.028^{*}$ & $+0.036^{*}$ \\
Applicability $J$ & $-0.011$ & $+0.025^{*}$ & $-0.034^{*}$ \\
Conflation gap (pp) & $+2.6^{*}$ & $+1.5$ & $-0.9$ \\
Same answer, \ins{e1} & 0.931 & 0.920 & 0.928 \\
Same answer, conformity & 0.937 & 0.937 & 0.937 \\
Same answer, applicability & 0.965 & 0.948 & 0.889 \\
\midrule
\multicolumn{4}{@{}l}{Gemma-4B (rationale) minus Gemma-4B} \\
Metric & RoT & CVAL & Country \\
\midrule
e1 accuracy & $+0.040^{*}$ & $+0.019^{*}$ & $+0.055^{*}$ \\
e1 Neutral recall & $+0.129^{*}$ & $-0.033^{*}$ & $-0.040^{*}$ \\
Conformity $J$ & $+0.020^{*}$ & $+0.060^{*}$ & $+0.157^{*}$ \\
Applicability $J$ & $+0.187^{*}$ & $+0.042$ & $+0.047^{*}$ \\
Conflation gap (pp) & $-40.8^{*}$ & $-24.0^{*}$ & $-9.0^{*}$ \\
Same answer, \ins{e1} & 0.870 & 0.864 & 0.852 \\
Same answer, conformity & 0.927 & 0.871 & 0.878 \\
Same answer, applicability & 0.797 & 0.582 & 0.516 \\
\bottomrule
\end{tabular}
\caption{Paired rationale-minus-no-rationale differences on identical items; $*$ = BH-significant. ``Same answer'' is the share of items given an identical answer by the two runs.}
\label{tab:reasoningcontrast}
\end{table*}

\begin{table*}[ht]
\centering
\small
\begin{tabular}{@{}lrrr@{}}
\toprule
Run & RoT & CVAL & Country \\
\midrule
\texttt{GPT-5.6} (option only) & $-0.4$ & $-1.9$ & $-4.1^{*}$ \\
\texttt{GPT-5.6} (rationale) & $+2.2^{*}$ & $-0.4$ & $-4.9^{*}$ \\
\texttt{Gemma-4B} (no rationale) & $+56.0^{*}$ & $+30.7^{*}$ & $+11.8^{*}$ \\
\texttt{Gemma-4B} (rationale) & $+15.2^{*}$ & $+6.7^{*}$ & $+2.8$ \\
\bottomrule
\end{tabular}
\caption{Conflation gap (pp) per run, same estimator as Table~\ref{tab:conflation}; $*$ = BH-significant over these 12 cells.}
\label{tab:reasoningconflation}
\end{table*}

A keyword reading of the written text (directional only, not annotation) shows the same split: Table~\ref{tab:reason} reports, at Country, how many gold-Neutral items each model answered Yes/No, how often its text hedges or uses softening versus firm-custom language, and how often a NotApply answer still mentions acceptability.

\begin{table*}[ht]
\centering
\small
\begin{tabular}{@{}>{\raggedright\arraybackslash}p{3.4cm}rr@{}}
\toprule
Measure (Country) & GPT-5.6 (rationale) & Gemini-3.8 \\
\midrule
Gold-Neutral forced to Yes/No & 777 & 797 \\
\quad of which the text still hedges & 169 (22\%) & 62 (8\%) \\
Softening language & 81\% & 17\% \\
Asserts a firm custom & 19\% & 63\% \\
NotApply answers with an acceptability word & 6 of 301 & 4 of 176 \\
Conformity $J$ & 0.60 & 0.88 \\
\bottomrule
\end{tabular}
\caption{Keyword readings of the written text at Country (directional only). ``Gold-Neutral forced to Yes/No'': gold-Neutral items answered Yes/No under \ins{e1} (of 797); the next row is the share of those whose text contains a hedge cue. ``Softening language'' and ``Asserts a firm custom'': share of all of the model's Country answers whose text contains a softening cue or a firm-custom cue (two keyword registers, not exhaustive). ``NotApply answers with an acceptability word'': among \ins{apply3strip\_verbose} Country NotApply answers, how many contain an acceptability word. The GPT-5.6 column is the rationale run; Gemini-3.8 writes a post-hoc explanation.}
\label{tab:reason}
\end{table*}

\section{Prompt specification}
\label{app:prompts}

Every prompt is assembled from the same scaffold,
\begin{quote}\small\sloppy
\textsc{task} + `` The answer options are '' + \textsc{letters} + ``.'' + [`` '' + \textsc{clause}] + \emph{newline newline} + \textsc{context} + \emph{newline newline} + ``Options:'' + \emph{newline} + \textsc{options} + \emph{newline} + ``Answer (options '' + \textsc{letters} + ``): ''
\end{quote}
where \textsc{letters} is ``A, B or C'' for the three-option instruments, ``A or B'' for \ins{binaryonly} and \ins{apply3strip\_verbose}, and ``A, B, C or D'' for \ins{joint4}. The context blocks are in Table~\ref{tab:context}, the base conformity prompt in Appendix~\ref{app:base}, and the change each instrument makes in Table~\ref{tab:promptmap}; the applicability and four-cell prompts, which replace more, are written out below.

\subsection{Context blocks}
Each level supplies a different context block (Table~\ref{tab:context}).
\begin{table}[ht]
\centering
\small
\begin{tabular}{@{}l>{\raggedright\arraybackslash}p{.72\columnwidth}@{}}
\toprule
Level & Context block \\
\midrule
RoT & Rule-of-thumb: \{\texttt{Rule-of-Thumb}\}\quad Story: \{\texttt{Story}\} \\
Country & Country: \{\texttt{Country}\}\quad Story: \{\texttt{Story}\} \\
CVAL & Country: \{\texttt{Country}\}\quad RoT: \{\texttt{Value}\}\quad Story: \{\texttt{Story}\} \\
\bottomrule
\end{tabular}
\caption{Context block per level. In CVAL the ``RoT:'' slot holds the dataset's abstract \texttt{Value}, not the Rule-of-Thumb field. \ins{apply3strip\_verbose} strips the dataset-native trailing acceptability question from the Story.}
\label{tab:context}
\end{table}

\subsection{Base conformity prompt}
\label{app:base}
The base task is shared by \ins{e1}, \ins{correcteddef}, \ins{full}, \ins{vague}, and \ins{binaryonly}, and the base option set is \texttt{A) Yes / B) No / C) Neither}. The task sentence changes only in the context sentence:
\begin{quote}\small\sloppy
\noindent\textbf{RoT.} You will be given a short story involving 2 or more characters about everyday situations. You will also be given a rule-of-thumb. Read the story and reason if the action by the character in the story is socially acceptable under the given rule-of-thumb alone. Do not make any extra inferences about actions outside of the given context and rule. Only align to the rule given.\\[4pt]
\noindent\textbf{Country.} You will be given a short story involving 2 or more characters about everyday situations. You will also be given a country of relevance. Read the story and reason if the action by the character in the story is socially acceptable in the given country. Do not make any extra inferences about actions outside of the given context and country. Only align to the country given.\\[4pt]
\noindent\textbf{CVAL.} You will be given a short story involving 2 or more characters about everyday situations. You will also be given a country of relevance, along with a rule-of-thumb. Read the story and reason if the action by the character in the story is socially acceptable in the given country with the associated rule-of-thumb. Do not make any extra inferences about actions outside of the given context.
\end{quote}
Two notes on the wording. In the CVAL context the dataset's abstract \texttt{Value} is supplied under the prompt's own ``rule-of-thumb'' label, so the CVAL sentence above is the benchmark's and is retained verbatim (Table~\ref{tab:context}). The base task says ``reason if the action $\ldots$ is socially acceptable'', but the answer itself is constrained to a single option letter or JSON field (Appendix~\ref{app:repro}), so any reasoning is internal to the model and is not emitted.

\subsection{What each instrument changes}
Table~\ref{tab:promptmap} lists, for every instrument, the base task it uses, the lines it adds or changes, the answer letters, and whether the story's trailing acceptability question is stripped.
\begin{table*}[ht]
\centering
\small
\setlength{\tabcolsep}{5pt}
\begin{tabular}{@{}ll>{\raggedright\arraybackslash}p{5.3cm}lc@{}}
\toprule
Instrument & Task & Added / changed lines & Letters & Story stripped \\
\midrule
\ins{e1} & conformity & (base) & A, B or C & no \\
\ins{correcteddef} & conformity & + column ``correcteddef'' of Table~\ref{tab:clauses} & A, B or C & no \\
\ins{full} & conformity & + column ``full'' of Table~\ref{tab:clauses} & A, B or C & no \\
\ins{vague} & conformity & + column ``vague'' of Table~\ref{tab:clauses} & A, B or C & no \\
\ins{binaryonly} & conformity & options reduced to \texttt{A) Yes / B) No} & A or B & no \\
\ins{apply3strip\_verbose} & applicability (below) & options replaced (below) & A or B & yes \\
\ins{joint4} & four-cell (below) & options replaced (below) & A, B, C or D & no \\
\bottomrule
\end{tabular}
\caption{Every change each instrument makes to the base prompt. Task and option text for the applicability and four-cell instruments are given below.}
\label{tab:promptmap}
\end{table*}

Table~\ref{tab:clauses} writes out the definition clause that each of \ins{correcteddef}, \ins{full}, and \ins{vague} appends at each level; \ins{vague} defines nothing and is the content-free control.
\begin{table*}[ht]
\centering
\small
\setlength{\tabcolsep}{3pt}
\begin{tabular}{@{}>{\raggedright\arraybackslash}p{2.4cm}>{\raggedright\arraybackslash}p{3.7cm}>{\raggedright\arraybackslash}p{3.7cm}>{\raggedright\arraybackslash}p{3.7cm}@{}}
\toprule
Clause & RoT & CVAL & Country \\
\midrule
\ins{correcteddef} & Answer C if the rule does not apply to the action in the story. & Answer C if the given country and rule-of-thumb do not apply to the action in the story. & Answer C if the given country's norms do not apply to the action in the story. \\
\ins{full} & Answer A if the action follows the rule-of-thumb. Answer B if the action violates the rule-of-thumb. Answer C if the rule does not apply to the action in the story. & Answer A if the action follows the given country and rule-of-thumb. Answer B if the action violates the given country and rule-of-thumb. Answer C if the given country and rule-of-thumb do not apply to the action in the story. & Answer A if the action follows the given country's norms. Answer B if the action violates the given country's norms. Answer C if the given country's norms do not apply to the action in the story. \\
\ins{vague} & \multicolumn{3}{>{\raggedright\arraybackslash}p{11.4cm}}{Answer C if the situation is neither clearly A nor clearly B, one way or the other. (identical at every level)} \\
\bottomrule
\end{tabular}
\caption{Definition clauses appended to the base task, by level. \ins{vague} does not define anything and is the content-free control.}
\label{tab:clauses}
\end{table*}

\subsection{Applicability prompt (\ins{apply3strip\_verbose})}
\begin{quote}\small\sloppy
\noindent\textbf{RoT task.} You will be given a short story involving 2 or more characters about everyday situations. You will also be given a rule-of-thumb. Decide only whether the rule-of-thumb bears on the action by the character in the story. Do not judge whether the action is acceptable.\\[4pt]
\noindent\textbf{Country task.} You will be given a short story involving 2 or more characters about everyday situations. You will also be given a country of relevance. Decide only whether that country's norms bear on the action by the character in the story. Do not judge whether the action is acceptable.\\[4pt]
\noindent\textbf{CVAL task.} You will be given a short story involving 2 or more characters about everyday situations. You will also be given a country of relevance, along with a rule-of-thumb. Decide only whether the given country and rule-of-thumb bear on the action by the character in the story. Do not judge whether the action is acceptable.\\[6pt]
\noindent\textbf{Options (RoT).} A) The rule-of-thumb bears on the action by the character in the story. \quad B) The rule-of-thumb does not bear on the action by the character in the story.\\[4pt]
\noindent\textbf{Options (Country).} A) That country's norms bear on the action by the character in the story. \quad B) That country's norms do not bear on the action by the character in the story.\\[4pt]
\noindent\textbf{Options (CVAL).} A) The given country and rule-of-thumb bear on the action by the character in the story. \quad B) The given country and rule-of-thumb do not bear on the action by the character in the story.
\end{quote}

\subsection{Four-cell prompt (\ins{joint4})}
The task is the base task's context sentence followed by: ``Read the story and decide which of the options below best describes the situation. Judge both whether \emph{the context} is relevant to the action, and whether the action itself seems socially acceptable, independent of \emph{it} if \emph{it} turns out not to be relevant,'' where \emph{the context} is the rule-of-thumb (RoT), the country's norms (Country), or the given country and rule-of-thumb (CVAL). The options are:
\begin{quote}\small\sloppy
A) \emph{The context} is relevant to this situation, and the action violates it.\\
B) \emph{The context} is relevant to this situation, and the action follows it.\\
C) \emph{The context} is not relevant to this situation, but independent of it, the action seems socially acceptable.\\
D) \emph{The context} is not relevant to this situation, but independent of it, the action seems socially unacceptable.
\end{quote}

\section{Reproducibility}
\label{app:repro}

We use the released \texttt{normad\_v2} data (2{,}615 items), each presented at three levels of context (RoT, CVAL, Country) and scored with the seven instruments above; the exact prompts are given in Appendix~\ref{app:prompts}. Every run is greedy (temperature $0$) and holds one variant per model, so a given item is always seen under the same prompt word for word.

The five open-weight models and the no-rationale Gemma run are served locally as BF16 GGUF weights through a \texttt{llama.cpp} OpenAI-compatible chat endpoint, with a GBNF grammar that restricts the completion to exactly one of the allowed answer letters (\ins{A}/\ins{B}/\ins{C}, \ins{A}/\ins{B}, or \ins{A}/\ins{B}/\ins{C}/\ins{D}); the recorded answer is that single letter, so no free-text parsing is needed. The two frontier models are called through a hosted API (the Kaggle Benchmarks API) instead: the answer is requested as JSON under a fixed schema, so every response is a single option field. Hidden chain-of-thought was disabled for the GPT-5.6 runs (confirmed in the call traces); Gemini-3.8 always reasons internally and never has it disabled. Prompts are identical across models apart from the final response-format instruction.

The two reasoning runs differ from their counterparts only by the added field. The GPT-5.6 rationale run uses the same API and JSON schema with a reasoning field written before the option, while its option-only counterpart returns the option field alone. The Gemma-4B rationale run uses the same BF16 weights as its no-rationale pair but a \texttt{transformers} stack with \texttt{outlines} JSON decoding of a \texttt{\{"reasoning", "answer"\}} object. \texttt{gemini-3.8-flash} emits hidden reasoning tokens that cannot be disabled and returns its explanation after the answer, so its text is a post-hoc explanation and it has no no-reasoning pair.

The NormAd data is already public; we will release the repository with the analysis code and the model run archives upon acceptance.

\end{document}